\documentclass[letterpaper, 10 pt, conference]{ieeeconf}  

\IEEEoverridecommandlockouts                              

\usepackage{graphicx} 
\usepackage{amsmath} 
\usepackage{amssymb}  

\title{\LARGE \bf
Multi-Instance Aware Localization for End-to-End Imitation Learning
}

\author{Sagar Gubbi Venkatesh$^{1, 2}$, Raviteja Upadrashta$^{2}$, Shishir Kolathaya$^{2}$, and Bharadwaj Amrutur$^{1, 2}$
\thanks{*This work was supported in part by Yaskawa India and Robert Bosch Center for Cyber-Physical Systems.}
\thanks{$^{1}$Department of Electrical and Communication Engineering, Indian Institute of Science, Bangalore 560012, India {\tt\small sagar@iisc.ac.in}; {\tt\small amrutur@iisc.ac.in}}%
\thanks{$^{2}$Robert Bosch Center for Cyber-Physical Systems, Indian Institute of Science, Bangalore 560012, India {\tt\small ravitejaupadras@iisc.ac.in}; {\tt\small shishirk@iisc.ac.in}}%
}

\begin{document}

\maketitle
\thispagestyle{empty}
\pagestyle{empty}

\begin{abstract}
Existing architectures for imitation learning using image-to-action policy networks perform poorly when presented with an input image containing multiple instances of the object of interest, especially when the number of expert demonstrations available for training are limited. We show that end-to-end policy networks can be trained in a sample efficient manner by (a) appending the feature map output of the vision layers with an embedding that can indicate instance preference or take advantage of an implicit preference present in the expert demonstrations, and (b) employing an autoregressive action generator network for the control layers. The proposed architecture for localization has improved accuracy and sample efficiency and can generalize to the presence of more instances of objects than seen during training. When used for end-to-end imitation learning to perform reach, push, and pick-and-place tasks on a real robot, training is achieved with as few as 15 expert demonstrations. 
\end{abstract}

\section{INTRODUCTION}\label{sec:intro}
There has been a lot of interest in recent times towards teaching robots to develop visuomotor skills \cite{end_to_end_deep_visuo_motor_policies}, \cite{compound_task_daml}, \cite{osiml}, \cite{daml}, \cite{vrteleop}. This paper attempts to address the challenge of using imitation learning to enable robots to perform manipulation tasks in the presence of multiple instances of an object in the scene. This is useful in automating applications such as garbage segregation or sorting operations performed in warehouses.

The underlying principle in imitation learning is to record observations seen and the actions taken by an expert when performing the task and to then train a policy network to clone the behavior of the expert. Imitation learning has been successfully applied to diverse problems ranging from self-driving cars \cite{nvidia} and drone navigation \cite{drone} to manipulation tasks with a robotic arm \cite{vrteleop}, \cite{daml}, \cite{yaskawa}, \cite{mdn}, \cite{siamese}.

Imitation learning requires high-quality expert demonstrations. While for some tasks such as navigation, it may be possible to instrument the car and record data non-intrusively in the background as the expert drives the car, recording data can be time-consuming and challenging for complex manipulation tasks with a robotic arm. Thus, it is highly desirable to choose network architectures with the appropriate inductive bias so as to improve sample efficiency. 

\begin{figure}[!t]
      \centering
      \includegraphics[width=0.95\linewidth]{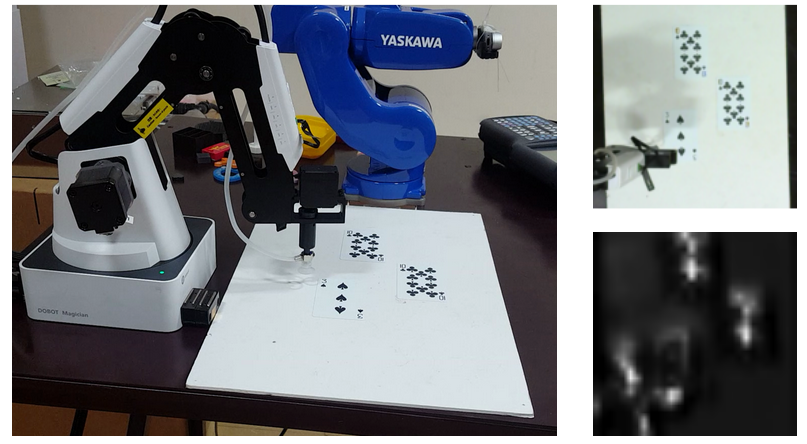}
      \caption{The robot must pick up the ten of clubs that is farthest from the camera. The heat map is from an intermediate layer in the end-to-end trained neural network and shows the network localizing the objects of interest even though no explicit supervision was given about the location of the object during training.}
      \label{fig:intro}
\end{figure}

In this paper, we consider end-to-end imitation learning to reach, push, and pick-and-place in the presence of multiple instances of the object of interest. The work presented in \cite{end_to_end_deep_visuo_motor_policies, vrteleop, osiml, daml} address similar objectives. The Convolutional Neural Network (CNN) architectures presented in these papers are similar and each one can be decomposed into, (a) vision layers, and (b) control layers. The vision layers are comprised of a few convolution layers followed by a spatial softmax layer. The spatial softmax layer enables them to attend to an object and to localize it while allowing sample efficient training. However, an inherent problem with spatial attention for localization is that it works reliably only when there is one instance of the object of interest. When there are multiple instances of the object, the output of the spatial softmax will be the weighted average of the locations of these objects, rather than the location of any of the objects in the scene. Thus, these architectures cannot be directly employed to perform tasks that involve multiple instances of objects.

Clearly, there has to be a mechanism by which the network is able to learn to focus on one particular instance of an object. We propose using an attention mechanism with a sense of ``instance awareness" embedded in it. With this, the policy network is capable of attending to one instance of the object while ignoring the rest.

Our contributions are:
\begin{itemize}
    \item We propose a network architecture for manipulation tasks with a robot arm in the presence of multiple instances of the object of interest.
    \item We evaluate the impact of the choice of network architecture on localization sample efficiency.
\end{itemize}


Expert data is collected via a teleoperation system to operate the Yaskawa MotoMini and the Dobot Magician robot arms (Figure~\ref{fig:intro}) using a keyboard. The neural network is trained for end-to-end imitation learning without any additional supervision or pre-training. Even though the action space is discrete in our setup, we find that object localization emerges as the network is trained to imitate the expert (Figure~\ref{fig:intro}). We compare the localization performance of a number of different architectures and show that the proposed architecture has better sample efficiency.



\section{RELATED WORK}

One way to design a robot controller for visual manipulation tasks such as pick-and-place is to construct a pipeline of separately trained modules \cite{amazon_picking_challenge}. For example, one might first train an object detector such as \cite{faster_rcnn} to detect all objects in the scene. The bounding box output (in 2D pixel coordinates) of such a detector can then be used by another network that is trained separately to predict the joint torques, which can be used to pick up the relevant object. In contrast to the pipelined approach, an end-to-end approach directly maps pixels to actions. This has the benefit of a richer, more flexible, and learnable intermediate representation that can offer better overall performance. The superior performance of the end-to-end approach compared to the pipelined approach has been observed on a number of problems ranging from scene text detection \cite{fsns} to speech recognition \cite{asr} and has shown promise in learning visuomotor policies \cite{vrteleop}.

Object detectors such as \cite{faster_rcnn} predict a large number of bounding boxes which are then suppressed using non-maximum suppression to obtain the final bounding boxes corresponding to objects in the scene. Since non-maximum suppression is not differentiable, it cannot be used for localization in end-to-end learning.

Fully Connected (FC) layers have been used to directly regress the position of objects from the feature map corresponding to the image \cite{sermanet2013overfeat}. Similarly, convolution layers have been used to regress the position of objects directly from the feature map \cite{deeppose}. Although both these methods are differentiable, they are not sample efficient. Spatial attention using the spatial softmax operation offers a sample efficient and differentiable way to localize objects.

\subsection{Spatial Attention for Localization}

The authors in \cite{vrteleop} show the effectiveness of imitation learning for training visuomotor policies using a small number of expert demonstrations collected using a virtual reality teleoperation system that interfaces with a PR2 robot. The end-to-end network in this paper maps raw RGBD images to angular and linear velocities of the robot joints. The input image is first processed by the vision layers, which consists of a few convolution layers, to obtain a feature map. The feature map is then passed through a spatial soft-argmax layer \cite{end_to_end_deep_visuo_motor_policies}, which serves as a spatial attention mechanism, that enables the network to attend to a region in the feature map corresponding to the object of interest. The output of spatial soft-argmax layer is then passed through the control layers of the network, which consists of a series of FC layers, to predict the robot actions. 

The utility of using the soft-argmax layer for developing deep visuomotor policies has been demonstrated in other papers as well. The work presented in \cite{osiml} combines imitation learning and model agnostic meta-learning \cite{maml} to infer a policy from a single human demonstration. The paper addresses the challenge of applying imitation learning in the context of one-shot learning. The work presented in \cite{daml} extends this algorithm for handling domain shifts when recording expert demonstrations. The authors of \cite{siamese} address the problem of one-shot localization using a siamese network with attention.

In \cite{kim2019attentive}, a distribution over functions is learnt to enable a neural network to predict about multiple functions consistent with the training data. In \cite{seker2019conditional}, a Gaussian distribution is predicted so that the network can handle the presence of multiple modes in the training data. Unlike these works, the training data is deterministic in our case. As mentioned earlier in Section \ref{sec:intro}, the drawback of using the spatial soft-argmax layer is that it can reliably localize the object of interest only when there is one instance of it in the scene. When multiple instances of an object are present in the scene, the network must be able to focus on one instance of the object while ignoring the others. For this, the network has to have a sense of ``instance awareness". 

\subsection{Instance Awareness}
The authors of \cite{fsns} develop an attention-based Recurrent Neural Network (RNN) architecture for recognizing relevant portions of text from an image of the scene. First, a feature map is extracted from the image using a CNN. The feature map is then spatially weighted using an attention mechanism, and the resulting output is fed to the RNN. The RNN outputs individual characters from relevant portions of the text present in the scene. This allows the architecture to learn to shift its focus from one location to another within the feature map over different time steps, thus allowing it to read multi-line text as well. 

The authors of \cite{uber_coord_conv} show that convolution fails at tasks that involve a coordinate transform problem (i.e. predicting pixel activation given spatial coordinates or vice versa). The authors propose a simple fix which involves appending the input to a standard convolution layer with 2 additional channels that contains spatial coordinate information. They call this the CoordConv layer. This enables networks to generalize and allow them to learn convolution filters that need to have spatial dependence embedded in them. The authors demonstrate the usefulness of this approach in image classification, object detection, generative modeling, and reinforcement learning. An attempt to use positional encoding combined with attention to improve performance in machine translation was made in \cite{attention_is_all_you_need} albeit with limited success.

\section{NETWORK ARCHITECTURE}

\begin{figure*}[!t]
      \centering
      \includegraphics[width=0.9\linewidth]{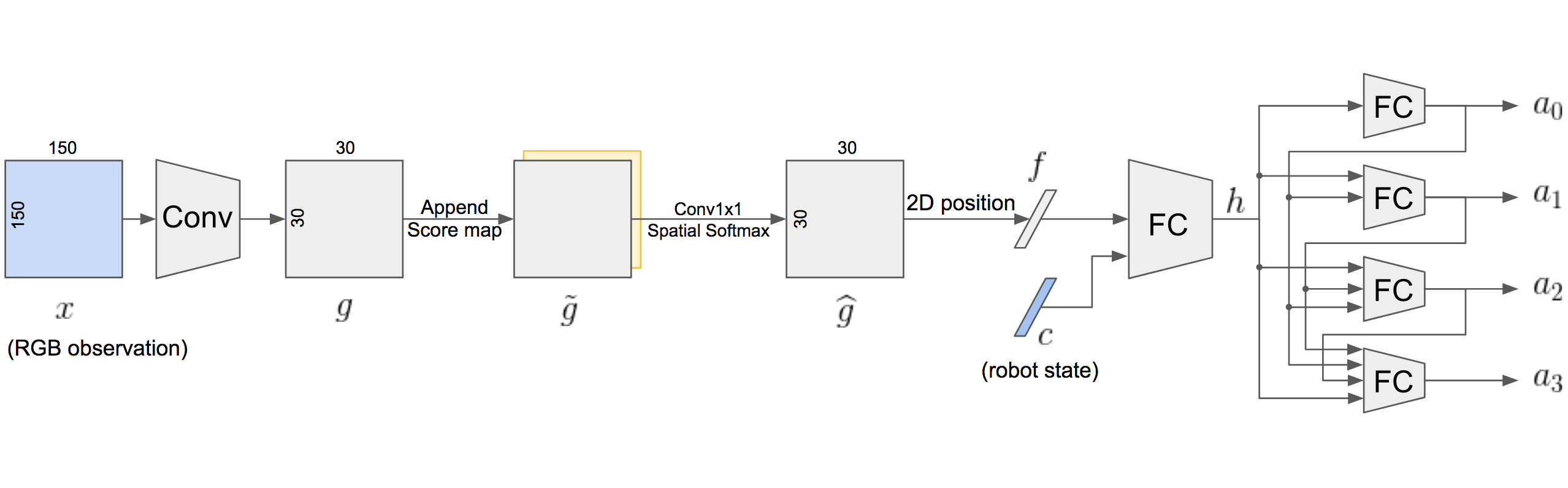}
      \caption{The proposed architecture. The convolutional (Conv) layers used are Conv3x3(16)-ELU-Conv3x3(32)-ELU-Conv3x3(64)-ELU-MaxPool2D-Conv3x3(64)-ELU-Conv3x3(128)-ELU-MaxPool2D-Conv3x3(128)-ELU-Conv3x3(128)-ELU-Dropout-Conv1x1(1)-Tanh. The intermediate FC layers are FC16-ReLU-FC7-ReLU. The output layers for $a_0$ to $a_2$ are FC8-ReLU-FC3-Softmax, and the output layer for $a_3$ is FC8-ReLU-FC2-Softmax. The feature map $g$ has only channel in our setup, but it can have more channels without any change in the subsequent layers. The use of pooling layers or strided convolutions in the ``Conv" block causes a reduction in the spatial resolution of $g$ compared $x$, but this also reduces the memory requirement and the number of learnable parameters. In our setup, the 30$\times$30 feature map can resolve to within 1.5~cm on the table. If the training data exhibits multi-modality, it may be desirable to use $h_t$ and $h_{t-1}$ to predict $a_t$.}
      \label{fig:arch}
\end{figure*}

The proposed neural network for end-to-end imitation learning is shown in Figure~\ref{fig:arch}. The RGB image from the camera $\mathbf{x}$ and the current state of the robot which includes the 3D pose of the end effector and the gripper state $\mathbf{c}$ are given as inputs to the network. The network predicts discrete actions $a_0$, $a_1$, $a_2$, and $a_3$. The first three actions correspond to the relative motion of the end effector along the X, Y, and Z directions. Each of these actions are discretized to 3 possible values: $\{-1, 0, +1\}$. The last action $a_3$ is binary and corresponds to the gripper open/close command.

The RGB image observation $\mathbf{x}$ is passed through several convolutional and pooling layers to produce the feature map $\mathbf{g}$. The intermediate value $\mathbf{f}$ corresponding to the 2D position of the points of interest in the image is obtained from $\mathbf{g}$ using layers described in more detail in the following paragraphs. The current state of the robot $\mathbf{c}$ and the 2D position of points of interest in the image, $\mathbf{f}$, are passed through several hidden layers to predict the actions $\mathbf{a}$. Rather than directly modeling the joint action space $p(\mathbf{a}|\mathbf{x}, \mathbf{c})$, we decompose it as shown in Eqn.~\ref{eqn:pixelcnn} to autoregressively predict the actions. This is preferable to assuming that the actions are independent, especially when there is stochasticity in the training dataset.

\begin{equation}
    p(\mathbf{a}|\mathbf{x}, \mathbf{c}) = p(a_0 | \mathbf{x}, \mathbf{c}) \prod_{i=1}^{3} p(a_i | a_0, a_1, ..., a_{i-1}, \mathbf{x}, \mathbf{c})
    \label{eqn:pixelcnn}
\end{equation}

We shall now consider the different ways one might obtain $\mathbf{f}$ from the feature map $\mathbf{g}$. Note that in all of these cases, no additional supervision is provided to learn a particular representation for $\mathbf{f}$, and the mapping from $\mathbf{g}$ to $\mathbf{f}$ is learnt as part of the end-to-end training under the constraint of the architecture we choose.

\subsection{Fully Connected Layers}
\label{sec:fc_layers}
The 2D position of the points of interest $\mathbf{f}$ may be derived from the feature map $\mathbf{g}$ by passing through one or more FC layers \cite{sermanet2013overfeat}. There is considerable freedom in how the network chooses to represent the position of the points of interest in $\mathbf{f}$. However, as we shall show in the next section, FC layers tend to overfit and generalize poorly when the training dataset is small as is often the case for imitation learning. 

\subsection{Convolutional Layers}
\label{sec:conv_layers}
The vector $\mathbf{f}$ may be obtained from $\mathbf{g}$ by passing it through several convolutional layers \cite{deeppose}. Unlike with FC layers, the spatial invariance of convolution acts as a stronger constraint resulting in improved generalization since the same operations applied to one spatial region of the feature map $\mathbf{g}$ have to be applied at all regions. However, as we show in the results, convolutional layers can also overfit and perform poorly.

\subsection{Spatial Softmax Layer}
\label{sec:spatial_softmax_layer}

\begin{figure}[!t]
      \centering
      \includegraphics[width=0.95\linewidth]{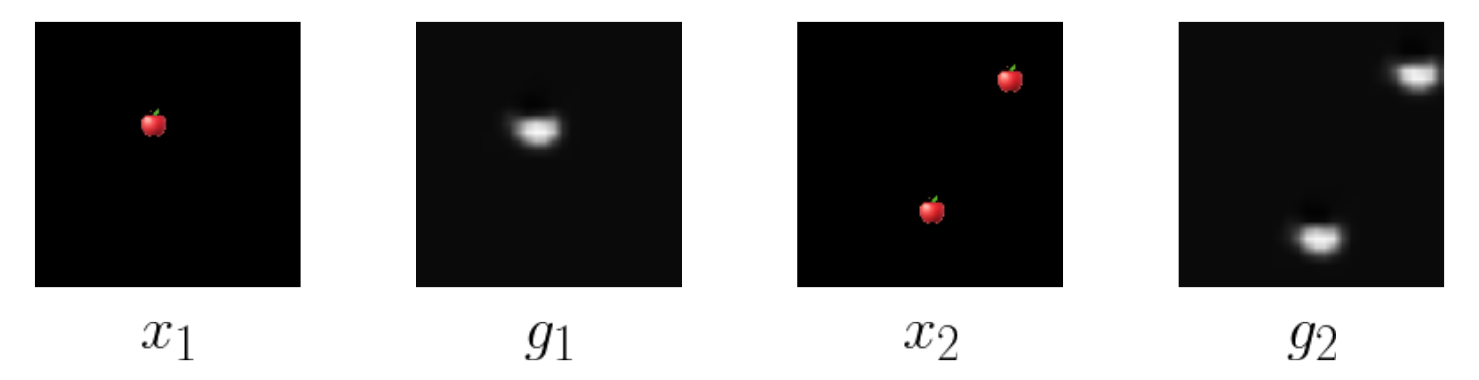}
      \caption{Feature map object when one and two instances of the object are present in the input image respectively.}
      \label{fig:one_two_heatmap}
\end{figure}

The 2D points of interest in the image may be obtained from the feature map $\mathbf{g}$ using the spatial softmax function.

\begin{equation}
    \widehat{g}^k_{i, j} = softmax_{i, j} \big(g^k_{i, j}\big)
    \label{norm_feature_map_eqn}
\end{equation}

\begin{equation}
    f^k_x = \sum_{i, j} \widehat{g}^k_{i, j} i
    \label{softargmax_x_eqn}
\end{equation}
\begin{equation}
    f^k_y = \sum_{i, j} \widehat{g}^k_{i, j} j
    \label{softargmax_y_eqn}
\end{equation}

where $1 \leq k \leq C_g$, with $C_g$ being the number of channels in $g$ and $0 \leq i < W_g$ and $0 \leq j < H_g$, with $W_g$ and $H_g$ being the width and height of $g$. This way of obtaining $\mathbf{f}$ has no learnable parameters at all. By construction, it generalizes well. To see why consider what happens when the object of interest is in the center of the image (see $x_1$ and $g_1$ in Figure~\ref{fig:one_two_heatmap}). Because a convolutional network is used to produce the feature map $\mathbf{g}$, due to spatial invariance, if the object is shifted in the image $\mathbf{x}$, the peak corresponding to the object in the feature map $\mathbf{g}$ is also shifted, and the spatial softmax will, by construction, give the correct localization in $\mathbf{f}$.

While this method of extracting position from the feature map generalizes well, it has a major shortcoming. It assumes that one and only one object of interest is present in the image. When multiple instances of the object of interest are present in the image, the feature map has multiple peaks each corresponding to one instance (see $x_2$ and $g_2$ in Figure~\ref{fig:one_two_heatmap}). When multiple peaks are present in the feature map $\mathbf{g}$ each corresponding to a different instance of the object in the image, the spatial softmax function gives incorrect localization by providing a (weighted) average 2D position of the peaks. The 2D position thus obtained may not correspond to the position of any object in the scene.

\subsection{Spatial Softmax after Appending One-Hot Encoded Position}
\label{sec:onehot_layer}

\begin{figure}[!t]
      \centering
      \includegraphics[width=0.99\linewidth]{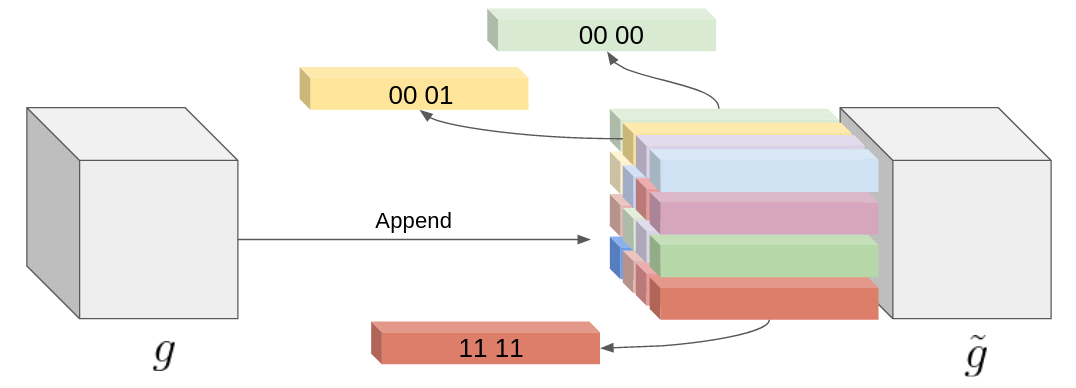}
      \caption{Illustration of appending one-hot encoding of position for a 4x4 feature map.}
      \label{fig:onehot_encoding}
\end{figure}

When there are multiple instances of the object of interest in the scene, say multiple apples, in order to pick one of the objects, it is necessary to specify an ordering of positions so that one position may be preferred over the other when the object of interest occurs at both positions. The preference for positions may be specified implicitly in the training data at the expense of sample complexity. One way to break the spatial invariance is to append the feature map $\mathbf{g}$ with a map where each pixel encodes its own position using a one-hot vector \cite{fsns} as shown in Figure~\ref{fig:onehot_encoding}. The augmented feature map $\mathbf{\tilde{g}}$ is passed through a 1$\times$1 bottleneck convolution and then the spatial softmax operation is performed as before. While the one-hot representation gives considerable flexibility, the notion of neighbourhood of pixels is lost. This results in poor generalization to new positions of the object that are not in the training set.

\subsection{Spatial Softmax after Appending 2D Coordinates}
\label{sec:coord_layer}

\begin{figure}[!t]
      \centering
      \includegraphics[width=0.99\linewidth]{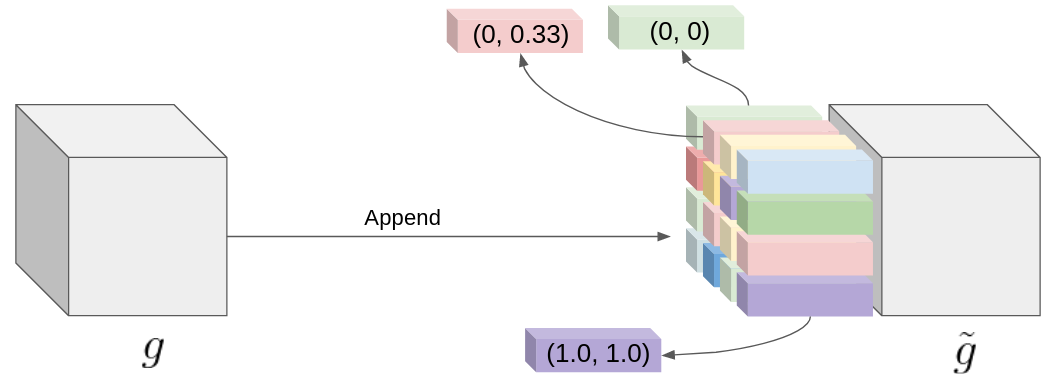}
      \caption{Illustration of appending 2D co-ordinate encoding of position for a 4x4 feature map.}
      \label{fig:coordinate_encoding}
\end{figure}

Similar to the previous idea, one might encode the position of each pixel of the feature map not by a one-hot vector, but by 2D coordinates $(x, y)$ as shown in Figure~\ref{fig:coordinate_encoding}. We would expect this representation to generalize better to positions unseen in the training set. However, the ordering of points indicating the preference of one position over another must still be specified by the data.

\subsection{Spatial Softmax after Appending Raster Scan Score Map}
\label{sec:score_layer}

\begin{figure}[!t]
      \centering
      \includegraphics[width=0.99\linewidth]{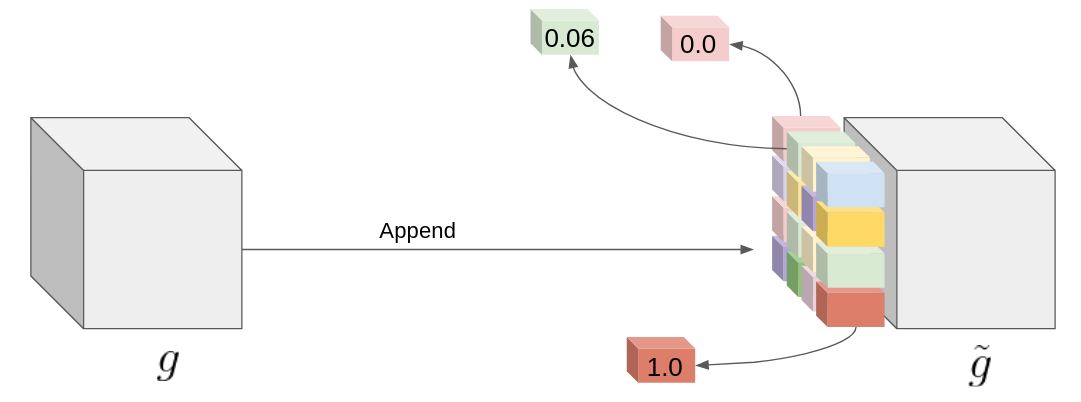}
      \caption{Illustration of appending score map to a 4x4 feature map. The score map is in raster scan order where the score corresponding to the first pixel of the feature map is 0.0, and the score corresponding to the last pixel is 1.0. }
      \label{fig:scoremap_encoding}
\end{figure}

In order to further reduce the number of training samples needed, it may be preferable to directly specify the order of preference of different positions. For example, we may want to pick objects from the scene in raster scan order. So, if the object of interest is present both in the left-top of the image and the right-bottom of the image, then we prefer the object at the left-top. To specify such an ordering directly rather than implicitly in the training data, we propose appending scores corresponding to the raster scan order to the feature map $\mathbf{g}$ as shown in Figure~\ref{fig:scoremap_encoding}. Note that the score map is just a linear function of the $(x, y)$ co-ordinates discussed in Section~\ref{sec:coord_layer} but with the weights frozen so that a particular ordering is enforced. The augmented feature map $\mathbf{\tilde{g}}$ is processed as before.

\section {RESULTS}

We first evaluate the different architectures discussed in the previous section on synthetic data for their localization performance. The proposed architecture is then evaluated for end-to-end imitation learning of manipulation tasks including reach, push, and pick-and-place.

\subsection{Localization Performance}

\begin{figure}[!t]
      \centering
      \includegraphics[width=0.9\linewidth]{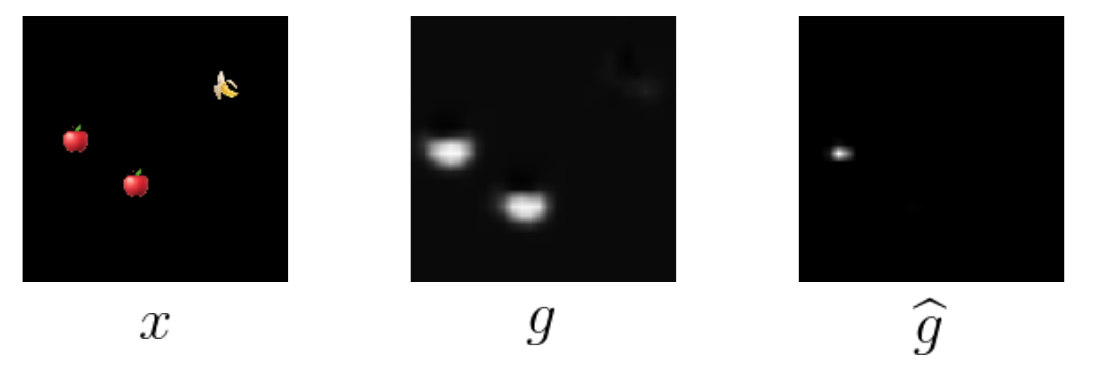}
      \caption{Feature map produced by the proposed architecture on an example from the synthetic dataset}
      \label{fig:apple_heatmap}
\end{figure}

We construct a synthetic dataset by placing apple emojis at random locations and distracting emojis at non-overlapping positions against a blank background as shown in Figure~\ref{fig:apple_heatmap}. The position of the first apple emoji (in raster scan order) is the label. In the training dataset, either one or two apple emojis are placed in three quadrants (all except the bottom-right quadrant), but in the test dataset up to three apple emojis and two distracting emojis are placed at random locations across the entire image. When measuring accuracy, we consider the output of the network to be a match if the pixel difference between the ground truth center and the predicted center of the object along both the axes is less than 8~px. Since we are only performing localization, the neural network architecture used ends at $\mathbf{f}$ (Figure~\ref{fig:arch}), and the subsequent layers are not used. 



\begin{figure}[!t]
      \centering
      \includegraphics[width=0.9\linewidth]{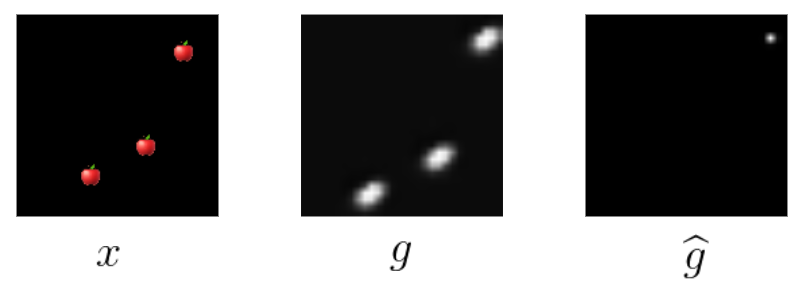}
      \caption{Feature map produced by the proposed architecture when the input image has 3 instances of the object. Note that during training, the network has been exposed only to images containing one or two instances of the object. }
      \label{fig:apple3_heatmap}
\end{figure}

 Table~\ref{table:loc11} compares the performance of the different architectures on a synthetic test dataset of 4096 samples. When using FC layers for localization (Section~\ref{sec:fc_layers}), the feature map $\mathbf{g}$ is obtained from the RGB image using the architecture in Figure~\ref{fig:arch} and is mapped to the 2D position $\mathbf{f}$ using FC layers\footnote{\label{note_fc}FC layers used are FC1024-ELU-FC128-ELU-FC2.}. Similarly, when convolution layers are used for localization, the 2D position $\mathbf{f}$ is obtained from $\mathbf{g}$ using convolution layers\footnote{\label{note_conv}Convolution layers used are Conv3x3-ELU-Conv3x3-ELU-Conv3x3-ELU-MaxPool2-Conv3x3-ELU-Conv3x3-ELU-MaxPool2-Conv3x3-ELU-Conv3x3-ELU-Conv1x1-ELU-Conv1x1.}. The other rows in Table~\ref{table:loc11} correspond to the methods described in Sections~\ref{sec:spatial_softmax_layer}~to~\ref{sec:score_layer}.
 
 Figure~\ref{fig:apple_heatmap} shows a sample feature map produced by the proposed architecture. We see that there are two peaks in the feature map $\mathbf{g}$ corresponding to the two apples in $x$. Following the proposed approach of appending a score map, we see that the peak corresponding to the apple at the bottom has been suppressed in $\widehat{g}$. Although the training set contains a maximum of two apples in the image, the proposed architecture generalizes and works even in the presence of three apples (Figure~\ref{fig:apple3_heatmap}). 
 
With the exception of the proposed architecture, the localization performance drops significantly when the training dataset size is reduced from 4096 to 32. Even though a dataset size of 32 may seem too small, the training samples are uncorrelated and generated at random, whereas it is common in imitation learning for successive observations to be highly correlated since they are successive frames of a demonstration. So, it is appropriate to evaluate performance on very small training sets.


\begin{table}[!t]
    \centering
    \begin{tabular}{p{3cm} p{2cm} p{2cm}} 
        \hline
        Method & Accuracy $N_{train}$ = 4096 & Accuracy $N_{train}$ = 32 \\ [0.5ex] 
        \hline
        FC layers & 55.71\% & 5.96\% \\
        Conv layers & 84.5\% & 13.04\% \\
        Spatial Softmax & 25.17\% & 23\% \\
        Spatial Softmax with one-hot encoded position & 99.51\% & 71.31\% \\
        Spatial Softmax with 2D co-ordinate map & 100\% & 91.5\% \\
        Spatial Softmax with raster scan score map (proposed) & 100\% & 99.39\% \\
        \hline
    \end{tabular}
    \caption{Localization accuracy in experiment A}
    \label{table:loc11}
\end{table}


\subsection{Imitation Learning Performance}

\begin{table}[!t]
    \centering
    \begin{tabular}{p{3cm} l l l} 
        \hline
        Method & Reach & Push & Pick-and-Place \\ [0.5ex] 
        \hline
        FC layers & 26.6\% & 13.3\% & 10\% \\
        Spatial Softmax with raster scan score map (proposed) & 100\% & 86.6\% & 85\% \\
        \hline
    \end{tabular}
    \caption{Task completion success rate in imitation learning}
    \label{table:imitation}
\end{table}

\begin{figure}[!t]
      \centering
      \includegraphics[width=0.9\linewidth]{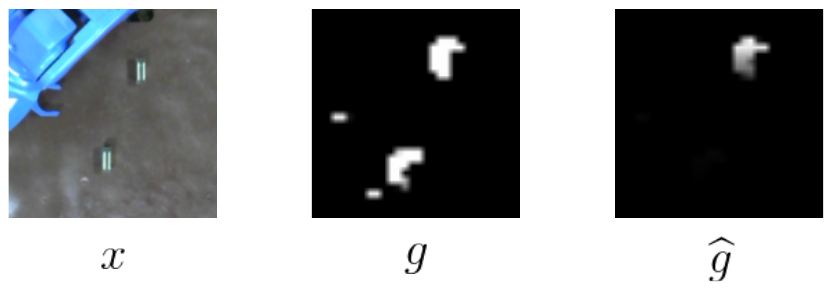}
      \caption{Feature map produced by the proposed architecture on a sample observation for the push task.}
      \label{fig:push_heatmap}
\end{figure}

\begin{figure}[!t]
      \centering
      \includegraphics[width=0.9\linewidth]{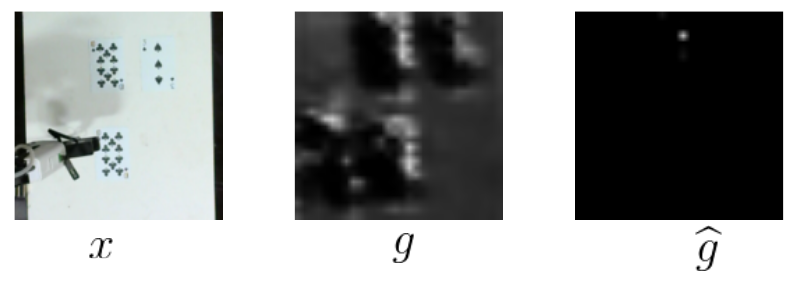}
      \caption{Feature map produced by the proposed architecture on a sample observation for the pick-and-place task.}
      \label{fig:pnp_heatmap}
\end{figure}

To evaluate the performance of the proposed architecture for imitation learning, we built a teleoperation system where the Yaskawa MotoMini and the Dobot Magician can be controlled through a keyboard. The RGB camera  along with the current state of the robot which includes the 3D end-effector pose and the gripper state are sampled and discrete motor commands are issued at 5~Hz. The network is trained using Adam optimizer \cite{adam}.

For the reach task, 15 demonstrations are collected for training and the learnt network is evaluated with 15 roll-outs. We consider the roll-out to be successful if the end-effector tip is within 2~cms of the target. For the push task, 15 expert roll-outs are collected for training and the trained network is evaluated with 15 roll-outs. The roll-out for pushing is considered successful if the object is pushed horizontally by at least 30~cms. For the pick-and-place task, we collected 10 expert demonstrations and evaluated the performance of the proposed network over 20 roll-outs. The pick-and-place is considered successful if the right playing card is picked up and placed at the edge of the table.

Table~\ref{table:imitation} compares the performance of the proposed approach with the FC layer approach for mapping $\mathbf{g}$ to $\mathbf{f}$ as described in Section~\ref{sec:fc_layers} which is equivalent to the architecture in \cite{vrteleop} with the spatial softmax layer removed due to its inability to fit the training data when multiple instances of the object are present in the scene. With the use of FC layers instead of the proposed architecture, we find that the network generalizes poorly and successfully completes the task only when the placement of objects is similar to a demonstration in the training set. For the proposed approach, the two failures in the push task were due to the object toppling over. With the pick-and-place task, in one instance, the suction failed to pick up the object and in the other two, the end-effector did not reach the right position to pick up the object. Figures~\ref{fig:push_heatmap}~and~\ref{fig:pnp_heatmap} show sample feature maps $\mathbf{g}$ and the corresponding input RGB images. Even though the action space is discrete, we find that localization of objects of interest has emerged as a consequence of training the network for end-to-end imitation learning. For the pick-and-place task, we find that the feature map $\mathbf{g}$ is more diffuse because of the size of the playing card happens to exceed the receptive field of the convolutional layers. As shown in Figure~\ref{fig:pnp_heatmap}, the network compensates for this by detecting the edge of the card in such a way that the subsequent layers can still localize the card.

\section{CONCLUSIONS}
Learning end-to-end visuomotor policies requires sample efficient and differentiable localization. We find that sample efficient localization in the presence of multiple instances of the object of interest in the scene requires breaking spatial invariance of convolution in a constrained manner. We propose doing this by appending a score map to the feature map produced by the convolution layers. The score map indicates our preference for one position over another when the object of interest is present at more than one location. When combined with spatial attention, the resulting localization layers are differentiable, sample efficient, and can work in the presence of multiple instances of the object of interest. The proposed network architecture generalizes to more instances of the object than in the training dataset and to locations not in the training dataset. When used for imitation learning to perform reach, push, and pick-and-place tasks, it enables learning from as few as 15 demonstrations.


\section*{ACKNOWLEDGMENT}
We thank Rokesh Laishram and Alok Rawat for assistance with the experimental setup.


\end{document}